\documentclass[conference]{IEEEtran}
\IEEEoverridecommandlockouts

\usepackage{cite}
\usepackage{amsmath,amssymb,amsfonts}
\usepackage{algorithmic}
\usepackage{graphicx}
\usepackage{textcomp}
\usepackage{xcolor}

\usepackage{microtype}
\usepackage{lipsum}
\usepackage{algorithm}
\usepackage{multirow}
\usepackage{caption}
\usepackage{subcaption}
\usepackage{url}

\usepackage{verbatim}
\usepackage[]{xcolor}

\usepackage{tabularx, ragged2e}

\newcommand{\specialcell}[2][c]{%
  \begin{tabular}[#1]{@{}c@{}}#2\end{tabular}}

\newcolumntype{C}{>{\Centering\arraybackslash}X}

\def\BibTeX{{\rm B\kern-.05em{\sc i\kern-.025em b}\kern-.08em
    T\kern-.1667em\lower.7ex\hbox{E}\kern-.125emX}}
\begin{document}

\title{Word Definitions from Large Language Models\\
}

\makeatletter
\newcommand{\linebreakand}{%
  \end{@IEEEauthorhalign}
  \hfill\mbox{}\par
  \mbox{}\hfill\begin{@IEEEauthorhalign}
}
\makeatother

\author{\IEEEauthorblockN{Bach Pham}
\IEEEauthorblockA{\textit{Department of Computer Science} \\
\textit{Earlham College}\\
Richmond, IN, USA \\
bqpham24@earlham.edu}
\and
\IEEEauthorblockN{JuiHsuan Wong}
\IEEEauthorblockA{\textit{Department of Computer Science} \\
\textit{Earlham College}\\
Richmond, IN, USA \\
swong21@earlham.edu}
\and
\IEEEauthorblockN{Samuel Kim}
\IEEEauthorblockA{\textit{Department of Computer Science} \\
\textit{Earlham College}\\
Richmond, IN, USA \\
skim24@earlham.edu}
\linebreakand
\IEEEauthorblockN{Yunting Yin}
\IEEEauthorblockA{\textit{Department of Computer Science} \\
\textit{Earlham College}\\
Richmond, IN, USA \\
yinyu@earlham.edu}
\and
\IEEEauthorblockN{Steven Skiena}
\IEEEauthorblockA{\textit{Department of Computer Science} \\
\textit{Stony Brook University}\\
Stony Brook, NY, USA \\
skiena@cs.stonybrook.edu}
}

\maketitle

\begin{abstract}
Dictionary definitions are historically the arbitrator of what words mean, but this primacy has come under threat by recent progress in NLP, including word embeddings and generative models like ChatGPT.
We present an exploratory study of the degree of alignment between word definitions from classical dictionaries and these newer computational artifacts.
Specifically, we compare definitions from three published dictionaries to those generated from variants of ChatGPT.
We show that (i) definitions from different traditional dictionaries exhibit more surface form similarity than do model-generated definitions, 
(ii) that the ChatGPT definitions are highly accurate, comparable to traditional dictionaries,  and
(iii) ChatGPT-based embedding definitions retain their accuracy even on low frequency words, much better than GloVE and FastText word embeddings.
\end{abstract}


\section{Introduction}
Many generations of readers have relied on the notion that the meaning of a word is what it says in the dictionary, but the primacy of dictionary definitions as the arbitrator of word meanings has come under threat by recent progress in natural language processing. Vector representations (word embeddings) have proven more valuable computationally than handcrafted semantics or definitions in describing the effective meaning of vocabulary words in a given language. And generative dialog systems like ChatGPT will happily produce full text definitions of every word when asked; indeed multiple versions of these definitions in response to different prompts. This paper address several distinct questions:
\begin{itemize}

    \item How consistent is the representation of semantics from classical dictionary definitions with those of word embeddings and generative language models?

    \item How quickly are the definitions generated by LLMs improving as the technology advances?
    
    \item LLMs have been described as ``stochastic parrots'' \cite{b1}.  To what extent are the definitions they produce mere repetitions or even plagiarisms of classical dictionary definitions?
\end{itemize}

For over 2,500 carefully selected words, we extract definitions from three different dictionaries (WordNet, Merriam-Webster, and Random House/Dictionary.com), and also vector representations from two prominent collections of word embeddings (Glove\cite{b2} and FastText\cite{b3}). We also asked two versions of ChatGPT (3.5 and 4.0) for the definitions of every word, each with two different prompts. We measure the distances between these representations using techniques including vector distance/similarity, edit distance, and neighborhood correlations.

Our primary observations include:

\begin{itemize}
    \item {\em The surface form of generated definitions differ substantially from those of published dictionaries} --
    Although ChatGPT presumably trained in part on the handcrafted definitions in our study, little-to-no unexpected trace of these texts remain in the generated text.

    \item {\em ChatGPT generated word definitions are highly accurate, consistent with published dictionaries} -- Of the 50 most distant pairs between Merriam-Webster and GPT4 definitions based on their definition embeddings, human evaluation confirms that GPT4 recognized 48 compared to 34 for GPT3, and that all but one of these definitions matched the primary sense of the published dictionary.

    \item {\em The consistency of LLM-generated definitions is relatively independent of frequency, unlike word embeddings} --
    We propose a new average distance correlation metric to compare word embeddings in different dimensional spaces. Our results suggest that SBERT embeddings of definitions (synthesized or published) may create more accurate word embeddings for low frequency words than traditional methods.

\end{itemize}

This paper is organized as follows.
Section \ref{sec:related-work} presents related work on LLMs and word embeddings.
The dictionaries and generative models we use to formulate our dataset are described in Section \ref{sec:dataset}.
The question of just how originial generated definitions are is considered in Section \ref{sec:longest_common_substr}.
We use vector representations to quantify the similarity of definitions by source in Section \ref{sec:distance_analysis},
and between definitions and word embeddings in Section \ref{sec:glove and fasttextword embeddings}.
We conclude with directions for future work in Section \ref{sec:conclusions}.

\section{Related Work}
\label{sec:related-work}
\subsection{Large Language Models}
Large language models (LLM) recently gained popularity, with ChatGPT reaching 100 million users in only three months \cite{b4}. Generative Pre-trained Transformer (GPT) models are a series of large language models developed by OpenAI. Each model in the series \cite{b5, b6, b7, b8} is trained on a larger corpus of text and achieves better results on natural language processing (NLP) tasks compared to its predecessors. LLMs have had significant impact across many domains, including medical imaging diagnostics \cite{b9}, recommendation systems \cite{b10}, and education \cite{b11}. They have been effectively applied to a variety of tasks such as taxonomy classification \cite{b12}, translation quality assessment \cite{b13}, and cloud configuration generation \cite{b14}.

\subsection{Word Embeddings}
Text embeddings provide a powerful representation of text data, and have been widely used across various applications, including bot detection \cite{b15}, mortality prediction \cite{b16, b17}, and other significant tasks. Word embeddings, in particular, represent words as multi dimensional vectors that encode both semantic and syntactic information. Words with similar meanings are positioned closer together in the vector space, making these embeddings highly effective for a range of NLP tasks. Prominent examples of word embeddings include Word2Vec \cite{b18}, GloVe \cite{b2}, and FastText \cite{b3}, which have been widely adopted in the NLP community.

\section{Dataset}
\label{sec:dataset}

Rate limitations on using the LLM APIs prevented us from studying the definitions of all English-language words, so we were selective in what words we study. Starting from a lexicon containing 246,591 English words sorted by their frequency, we sampled 3000 words by including the 1000 most frequent words and additional words drawn from four frequency tiers: 5000-5500, 10000-10500, 20000-20500, and 50000-50500. We cleaned the dataset by removing any word whose definition is missing in any source.
This leaves 2512 words with complete definitions remaining in our final dataset.

For each selected lexicon word, we collect answers from both GPT3.5 and GPT4 using two different prompts, identified throughout the text as type 1 or 2 prompts:
\begin{itemize}
    \item Type 1: ``What is the meaning of this word?"
    \item Type 2: ``Define this word."
\end{itemize}

This experimental design enables us to make a variety of comparisons between definitions produced (a) between different prompts on the same model, (b) two different models, namely GPT3.5 and GPT4.0, (c) comparisons between models and published dictionaries, (d) between common, moderate, and rare words, and (e) between different parts of speech. All the experiments in this paper are performed on a server with Intel Xeon Gold 6140 CPU (2.30GHz, 36 cores, 25MB Cache).

To provide fair grounds for comparison, we extracted published definitions for each of these words from three online dictionaries, namely: WordNet \cite{b19}, Merriam-Webster dictionary \footnote{\url{https://www.merriam-webster.com/}}, and Dictionary.com \footnote{\url{https://www.dictionary.com}}.

The definition length distribution of these sources is summarized in Table \ref{table:def length summary}, including mean and min/max lengths and standard deviation.
Observe that GPT4 definitions are generally a little longer than those produced by GPT3.5, but within a given model the two prompts produce similar length texts. The correlation of the lengths of the two GPT3 definitions is 0.61, and that of the two GPT4 definitions is 0.74. Two of the published dictionaries (Merriam-Webster and Dictionary.com) produce substantially longer definitions than WordNet and GPT3.5.

\begin{table}
\centering
\small
\begin{tabular}{|r|r|r|r|r|r|}
\hline
\textbf{\specialcell{Source}} & \textbf{N} & \textbf{\specialcell{Avg \\ Length}} & \textbf{\specialcell{Min}}  & \textbf{\specialcell{Max}} & \textbf{\specialcell{$\sigma$}} \\ \hline

\specialcell{GPT3-1} & 2459 & 35.97 & 3 & 253 & 20.21\\  
\specialcell{GPT3-2} & 2488 & 43.09 & 3 & 248 & 28.65\\  \hline
\specialcell{GPT4-1} & 2488 & 132.31 & 20 & 386 & 64.97\\  
\specialcell{GPT4-2} & 2489 & 133.77 & 28 & 466 & 65.26\\  \hline
\specialcell{Wordnet} & 2247 & 37.25 & 1 & 416 & 45.36\\  
\specialcell{Merriam-W.} & 2231 & 132.43 & 2 & 1957 & 179.59\\  
\specialcell{Dict.com} & 2398 & 223.11 & 3 & 3049 & 328.89\\  \hline

\end{tabular}
\caption{Definition length summary for collected dictionary datasets, showing the mean, min, max and standard deviation for each source, plus the number of queried lexicon terms defined within each sources (out of 3000).}
\label{table:def length summary}
\end{table}

\section{Does ChatGPT Plagiarize?}
\label{sec:longest_common_substr}

Although neural network-based generative models synthesize response texts instead of explicitly cutting-and-pasting from training texts, this does not mean they cannot plagiarize, say through overfitting model parameters.
Dictionary definitions provide an interesting domain to assess this possibility: they are precisely written texts that are readily available in training data. We investigate this issue by identifying the longest common substring between the ChatGPT-generated definition for a given word and the corresponding definition in a published dictionary.
Table \ref{LCS} reports the words which contain the longest case-insensitive match between each model/dictionary pair.
These examples are quite compelling, with common sequences as long as 17 words capturing the heart of the definition.

\begin{table*}
\centering
\begin{footnotesize}    
\begin{tabularx}{\textwidth}{|l|l|l|l}
\hline
\textbf{Source} & \textbf{GPT} & \textbf{Word} & \textbf{Longest Common Substring} \\ \hline 
\multirow{3}{*}{Wordnet} & 3-1 & clinic & a medical establishment run by a group of medical  \\ 
 & 3-2 & minutes & a unit of time equal to 60 seconds or 1/60th of an hour \\ 
 & 4-1 & forerunner & that precedes and indicates the approach of something or someone \\ 
 & 4-2 & nodule & ally harder than the surrounding rock or sediment \\
\hline
\multirow{3}{*}{\specialcell{Merriam\\-Webster}} & 3-1 & kelvin & the base unit of temperature in the International System of Units \\ 
 & 3-2 & town & ally larger than a village but smaller than a city \\
 & 4-1 & atheism & a philosophical or religious position characterized by disbelief in the existence of a god or any gods \\ 
 & 4-2 & econometric & tical methods to the study of economic data and problems \\ 
\hline
\multirow{3}{*}{\specialcell{Dictionary\\.com}} & 3-1 & letter &  addressed to a person or organization and usually transmitted by mail \\ 
 & 3-2 & delta & the fourth letter of the Greek alphabet ($\Delta$, $\delta$) \\ 
 & 4-1 & back & he rear part of the human body, extending from the neck to the lower end of the spine \\ 
 & 4-2 & compared & o examine (two or more objects, ideas, people, etc.) in order to note similarities and differences \\ 
\hline
\end{tabularx}
\end{footnotesize}
\caption{\label{LCS}
The longest common substring between GPT-model generated definitions and published dictionaries.   These matching phrases often capture the primary sense of the underlying word.
}
\end{table*}

Determining a meaningful expected longest common substring length in our instance is challenging, for several reasons.
For randomly generated sequences, statisticians have proven this quantity grows logarithmically in the length of the sequences \cite{b20, b21}.
But natural language text is far from random, and uses a large vocabulary as opposed to a constant-sized alphabet.
Further, text with a high semantic similarity (two definitions of the same word) should share greater surface similarity than mismatched definitions.
The length of the definitions matter, in a non-trivial way: longer, more detailed definitions should be expected to contain longer matches than briefer descriptions.

To assess whether these common phrases reflect untoward plagiarism or instead naturally constrained word choices in precise definitions, we compare the degree of borrowing between published dictionaries with what one sees generated by models.
Standards for acceptable borrowing in published dictionaries are established by social conventions and copyright law. Figure \ref{fig:freq_matched-summary} presents the frequency distributions of match length between all three pairs of published dictionaries, and the two prompts for each model.
They show that each pair of dictionaries share a small tail of long common substrings in their definitions, with substantially greater sharing between the two definitions from GPT-3 and GPT-4.
There is nothing untoward about this, just that the generative models repeat language in the two equivalent definitions.
Surprisingly, GPT-4 repeats itself substantially more frequently than GPT-3, perhaps an artifact from efforts to reduce model hallucination. The average length of common strings, and frequencies of long matches shows more aggressive borrowing between Merriam-Webster and Dictionary.com than any other pairing of published dictionaries or dictionary-model pairs.
In particular, 155 long matches of length $\ge 5$ are observed between Merriam-Webster and Dictionary.com, which is roughly twice as many as between either of these dictionaries and any model. There seems no real evidence that GPT models unfairly replicate published training data, despite the long matches reported in Table \ref{LCS}.

\begin{figure}
\centering
\begin{subfigure}{\linewidth}
  \centering
  \includegraphics[width=0.8\linewidth]{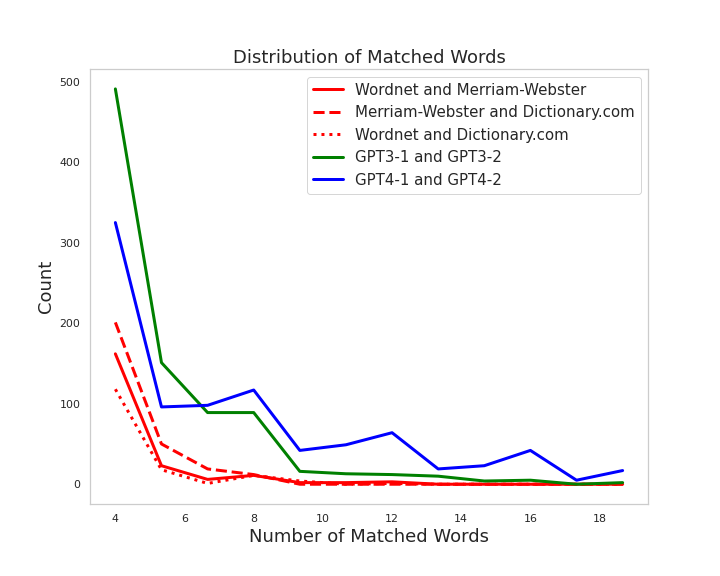}
  \caption{Match length distributions for dictionaries and GPT models.}
  \label{fig:matched_among_dict}
\end{subfigure}%
\\
\begin{subfigure}{\linewidth}
  \centering
  \includegraphics[width=0.8\linewidth]{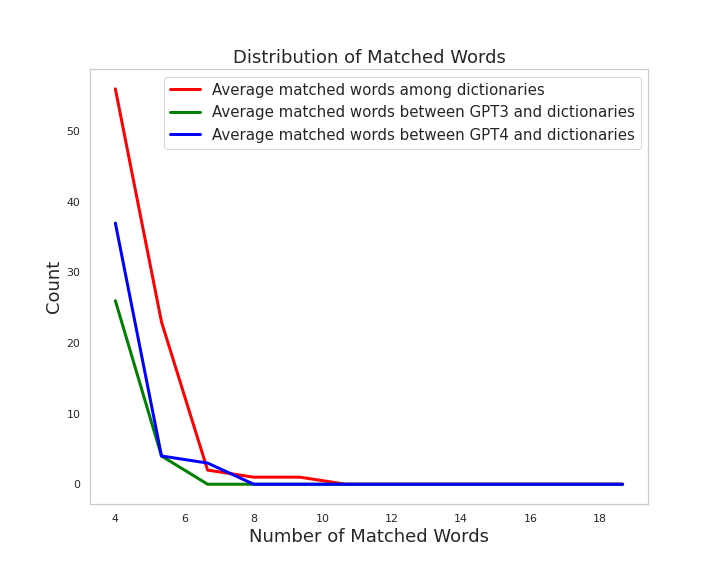}
  \caption{Average match length distributions against dictionaries.}
  \label{fig:matched_among_gpt}
\end{subfigure}
\caption{Frequency distribution of number of matched words, among dictionary pairs and GPT models (top), and the average number of matched words among dictionaries and between GPT model and dictionaries (bottom). GPT models exhibit substantially more borrowing than published dictionaries.}
\label{fig:freq_matched-summary}
\end{figure}

\section{Are GPT Definitions Consistent with Dictionaries?}
\label{sec:distance_analysis}

Model-generated word definitions aim to capture the same essential meaning as published dictionaries, while avoiding textual equivalence.
In order to obtain quantitative information about the relative quality of published and generated definitions, we consider three distinct distance or similarity functions between short texts---here definitions of the same word from different sources.
Two of these are based on the distance between embeddings, while the third works in the full text space of the definitions:

\begin{itemize}
    \item {\bf SBERT} \cite{b22} -- a modification of BERT that uses Siamese network structures to extract sentence-level representation using triplet loss.
    \item {\bf MPnet} \cite{b23} -- a pre-trained model that combines permuted language modeling and the use of auxiliary position information to derive feature embedding.
    \item {\bf Edit Distance} -- a string metric that represents the minimum number of insertions, deletions, or replacements required to transform one string into another. Here the edit distance is normalized by dividing the maximum length of the two strings.
\end{itemize}

We calculate the distance between definitions provided by GPT models and online dictionaries. The definitions from ChatGPT appear closer to Wordnet and Merriam-Webster, while definitions from GPT4 are closer to those on Dictionary.com. We use cosine distance of SBERT embeddings as our distance function in the rest of this paper, although similar results follow from MPNet.

\subsection{POS Analysis}
\label{sec:pos}

A natural hypothesis is that certain classes of words are easier to generate reliable definitions for than others.
Nouns represent objects, which might appear to be easier to precisely than descriptive words like adjectives and adverbs.
We obtain a dominant POS tag on each word using the NLTK library \cite{b24}. The average cosine distance error between generated and published definitions is calculated for each model and POS type. Curiously, GPT3.5 proves most accurate on descriptive words, which GPT4 produces its best definitions for nouns. The results are consistent across all three published dictionaries, and indistinguishable for Type 1 and Type 2 prompts.

\subsection{Word Frequency Analysis}
\label{sec:frequency}

There are natural but contradictory hypotheses to govern how the quality of LLM-generated definitions should vary as a function of the relative frequency of each word.
One may speculate that the most common words are hardest to define, because of functional forms like prepositions, and that they are more likely to be enriched with multiple senses.
But it is equally reasonable to think that low frequency words will be most difficult for generative models to understand, as they are seen least frequently in training data.

To resolve this debate, we compare the average cosine distance between SBERT-encoded dictionary and generated definitions, partitioned by class into words of high frequency, moderate frequency, and rare words.
The results in Table \ref{table:cosine_distance_gpt_dictionary_frequency} show words of middle frequency produce the best dictionary-generated for all models, across all dictionaries.
The differences between frequency tiers is generally quite modest, less than the impact of model version.
GPT-3.5 appears to generate slightly better definitions than the later GPT-4 by this metric, while the choice of prompt has little effect on the accuracy of result from either model.

\begin{table}
\centering
\small
\setlength{\tabcolsep}{5pt}
\begin{tabular}{|r|r|c|c|c|}
\hline
\textbf{Model}  & \textbf{Frequency} & \textbf{\specialcell{Word \\ Net}} & \textbf{\specialcell{Merriam-\\Webster}}  & \textbf{\specialcell{Dictionary.\\com}}\\ \hline
\multirow{3}{*}{GPT3-1} & high & 0.35 & 0.34 & 0.42 \\ 
& medium & \textbf{0.28} & \textbf{0.28} & 0.31 \\ 
& low & 0.32 & \textbf{0.28} & \textbf{0.30} \\ \hline
\multirow{3}{*}{GPT3-2} & high & 0.35 & 0.33 & 0.41 \\ 
& medium & \textbf{0.29} & \textbf{0.27} & \textbf{0.30} \\ 
& low & 0.34 & 0.29 & \textbf{0.30}\\ \hline
\multirow{3}{*}{GPT4-1} & high & 0.40 & 0.37 & 0.37 \\ 
& medium & \textbf{0.34} & \textbf{0.29} & \textbf{0.28} \\ 
& low & 0.37 & \textbf{0.29} & \textbf{0.28} \\ \hline
\multirow{3}{*}{GPT4-2} & high & 0.37 & 0.34 & 0.35 \\ 
& medium & \textbf{0.33} & \textbf{0.28} & \textbf{0.28} \\ 
& low & 0.39 & 0.30 & \textbf{0.28} \\ \hline

\end{tabular}
\caption{Cosine distance based on SBERT for different word frequency.}
\label{table:cosine_distance_gpt_dictionary_frequency}
\end{table}

\subsection{Are GPT-Generated Definitions Accurate?}
Researchers have found that ChatGPT can easily generate answers with complete assurance, even sometimes the answer is wrong \cite{b25}. To evaluate the correctness of GPT-generated definitions, we manually compare the definitions of the 50 words with the largest Euclidean distance between embedding definitions between Merriam-Webster and GPT-4.
Of these fifty words we manually evaluate, the biggest distances occur when models admit they do not know the definitions of generally obscure words (often proper names and abbreviations) that appeared in Merriam-Webster.
Only two of the fifty words (``cordier" and ``imon") were unknown to GPT-4, compared to 16 words GPT-3 did not understand.
Of the 48 words for which GPT-4 ventured a definition, we deemed all of them similar except for ``Acton'', which denoted a place in GPT-4 and a person in Merriam-Webster.
We conclude that the model generated definitions are generally of high quality, consistent with those from published dictionaries.

\section{Word Embeddings and Definitions}
\label{sec:glove and fasttextword embeddings}

Word embeddings are vector representations that capture the semantics of word usage.
What dictionary definitions are for people, word embeddings are for NLP models: an easily-understood representation of the meaning of a vocabulary word.
In this section we will directly compare traditional word embeddings to text embeddings of explicit dictionary embeddings, to help establish the level of correspondence between these representations.
Specifically, we obtain GloVE and 300-dimensional FastText word embeddings on all the words, and compare the closest neighbors of a word based on definition and word embeddings. 
We cannot compare these embeddings directly, because the underlying spaces and even dimensionality are incomparable.

Hence, we propose a new technique to measure the consistency of word and definition embeddings over a common vocabulary.
For each word $w$ in the lexicon, we calculate the distance from $w$ to all other words, separately in word embedding and definition space.
The correlation coefficent $r$ between these distances defines the agreement between the spaces from the perspective of word $w$.
To get a full vocabulary metric on space similarity, we average these correlations for each word in the lexicon.

Table \ref{table:corr_glovewordembed_defembed_cosine} compares published and generated definitions with popular word embeddings, GloVE and FastText, using the correlation measure defined above.
We distinguish between our three tranches of words, high-frequency, moderate, and rare.

The results are consistent across all sources and embeddings: high-frequency words show better consistency between definition-word embeddings than moderate-frequency words, which are substantially better than low frequency words.
As our results in Section \ref{sec:frequency} show the definitions of roughly equal quality across frequency tranches, the difference must be due to the word embeddings themselves becoming less accurate with diminishing usage frequency.
This presumably results from embeddings that are trained on insufficient amounts of data.

\begin{table}
\centering
\small
\setlength{\tabcolsep}{5pt}
\begin{tabular}{|r|r|r|r|}
\hline
\textbf{Model}  & \textbf{Frequency} & \textbf{Glove} & \textbf{FastText}\\ \hline
\multirow{3}{*}{GPT3-1} & high & \textbf{0.27} & \textbf{0.32} \\ 
& medium & 0.21 & 0.28 \\ 
& low & 0.08 & 0.22 \\ \hline
\multirow{3}{*}{GPT3-2} & high & \textbf{0.28} & \textbf{0.35} \\ 
& medium & 0.22 & 0.30\\ 
& low & 0.13 & 0.24 \\ \hline
\multirow{3}{*}{GPT4-1} & high & \textbf{0.28} & \textbf{0.32}\\ 
& medium & 0.23 & 0.30\\ 
& low & 0.11 & 0.25\\ \hline
\multirow{3}{*}{GPT4-2} & high & \textbf{0.30} & \textbf{0.38}\\ 
& medium & 0.26 & 0.35\\ 
& low & 0.12 & 0.29\\ \hline
\multirow{3}{*}{Wordnet} & high & \textbf{0.22} & \textbf{0.26} \\ 
& medium & 0.19 & 0.22\\ 
& low & 0.11 & 0.16\\ \hline
\multirow{3}{*}{Merriam-Webster} & high & \textbf{0.25} & \textbf{0.30}\\ 
& medium & 0.18 & 0.23\\ 
& low & 0.06 & 0.18\\ \hline
\multirow{3}{*}{Dictionary.com} & high & \textbf{0.33} & \textbf{0.31} \\ 
& medium & 0.21 & 0.25 \\ 
& low & 0.08 & 0.20 \\ \hline

\end{tabular}
\caption{Pearson correlation between word embedding and definition embedding based on cosine distance for different word frequency.}
\label{table:corr_glovewordembed_defembed_cosine}
\end{table}

\section{Conclusion and Future Work}
\label{sec:conclusions}
Our work demonstrates that while LLM-generated definitions differ from classical dictionary definitions in phrasing, they remain consistent and accurate in meaning. We also reveal that LLM-generated definitions are robust, particularly for low-frequency words, where they outperform traditional word embeddings. The scope of future work can be expanded to include multilingual definitions and domain-specific vocabularies to gain more insights into the potential and limitations of LLMs for semantic representation and language understanding.

\end{document}